\pdfoutput=1

\documentclass[11pt]{article}

\usepackage[preprint]{acl}

\usepackage{times}
\usepackage{latexsym}
\usepackage[T1]{fontenc}

\usepackage{xurl}

\usepackage[whole]{bxcjkjatype}

\usepackage[utf8]{inputenc}

\usepackage{microtype}

\usepackage{inconsolata}

\usepackage{graphicx}

\usepackage{multirow}

\title{YOMI-Bench: A Benchmark for Evaluating \\ Kanji Reading and Phonological Understanding of LLMs for Japanese}

\author{
Ryota Mibayashi$^{1}$,
Hiroya Takamura$^{2}$,
Hitomi Yanaka$^{3,4,5}$ \\[0.5em]
$^{1}$ Kobe University 
$^{2}$ National Institute of Advanced Industrial Science and Technology (AIST) \\
$^{3}$ The University of Tokyo
$^{4}$ RIKEN 
$^{5}$ Tohoku University \\
}

\begin{document}
\maketitle
\begin{abstract}
We propose YOMI-Bench, a benchmark for evaluating kanji reading and phonological understanding of large language models (LLMs) for Japanese.
In Japanese, a single kanji character often has multiple possible readings, making it difficult to infer the correct reading from surface-level text alone.
Due to these linguistic characteristics, it is empirically known that LLMs exhibit low performance in kanji reading for Japanese.
The proposed YOMI-Bench consists of four tasks specifically designed to evaluate kanji reading performance in Japanese.
In our evaluation using YOMI-Bench, we assessed one multilingual open LLM, four Japanese-specific open LLMs, and five commercial LLMs.
As a result, we found that even Japanese-specific models show low performance, and that commercial models also perform poorly on generation tasks that require consideration of kanji readings.
\end{abstract}

\section{Introduction}
The multilingual abilities of large language models (LLMs)  have been analyzed from various perspectives~\cite{zhu2024multilinguallargelanguagemodels}.
Among these, we focus on the ability of LLMs to recognize and utilize the phonological readings of text, which is related to tasks such as grapheme-to-phoneme (G2P) estimation and phoneme-aware text generation.
The relationship between written forms and their pronunciations varies substantially across languages, and even among languages that share the same writing system, notable differences can be observed.
For example, although Chinese and Japanese share many kanji characters, nearly all kanji characters in Chinese correspond to a single pronunciation, except for approximately 10\% of cases~\cite{Matsuo2010Neural}.
In contrast, approximately 60\% of Japanese kanji have multiple possible readings (see details in Sec.~\ref{sec:data})
, requiring more complex linguistic information for correct interpretation.
For instance, the kanji character ``覚'' has three possible readings: ``\underline{\textit{kaku}},'' ``\underline{\textit{obo}},'' and ``\underline{\textit{sa}}.''
These readings vary depending on the word in which the character appears, such as ``覚醒 (\textit{\underline{kaku}sei}/Awakening),'' ``覚える (\textit{\underline{obo}eru}/Memorize),'' and ``覚める (\textit{\underline{sa}meru}/Wake up).''
Thus, in order to correctly predict the readings of kanji characters appearing within individual words, models should not only possess the knowledge of the multiple readings of each kanji character, but also predict the correct reading based on the word.
For practical situations, the ability to correctly understand the readings of kanji is important for LLMs to solve classification and generation tasks that take rhyming into account, such as proofreading and generating lyrics~\cite{potash2015ghostwriter, nikolov2020conditional}, rap verses~\cite{xue2021deeprapper, mibayashi2023rapbattle}, and advertising texts~\cite{Lei2022plato}.

\begin{figure}[t]
  \includegraphics[width=\linewidth]{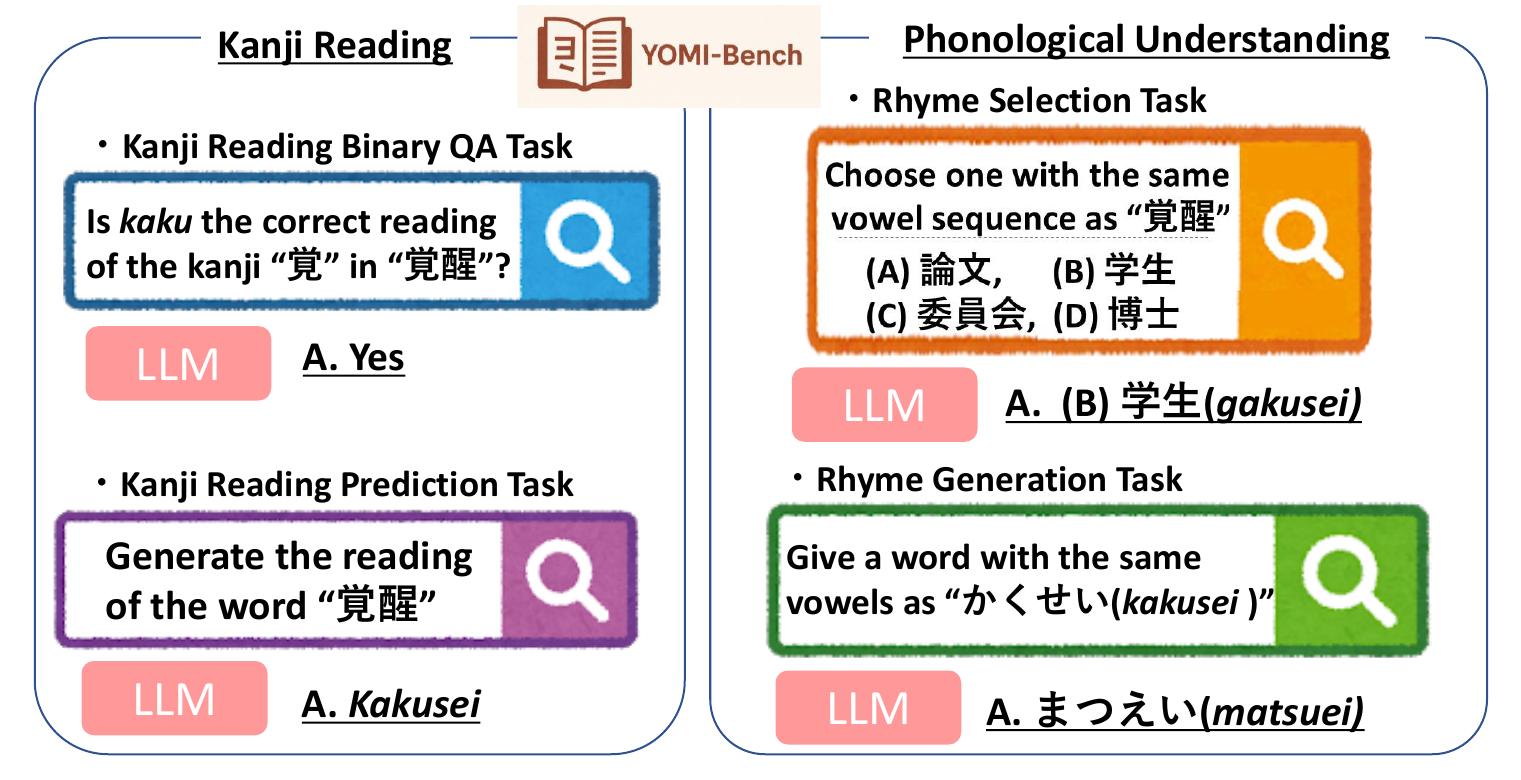}
  \caption{The overview of YOMI-Bench.}
  \label{fig:abst}
  \vspace{-1.0em}
\end{figure}

However, the reading ability of LLMs that considers such linguistic characteristics of Japanese remains largely unexplored.
Therefore, as illustrated in Figure~\ref{fig:abst}, we construct YOMI-Bench, a benchmark for evaluating Japanese reading performance of LLMs.
YOMI-Bench is a multi-task evaluation set involving seven types of binary/multiple QA and text generation tasks.
For example, the question ``Generate the reading of the word 覚醒'' asks LLMs to answer the correct kanji reading as a text generation task,  where the gold answer is ``\textit{kakusei}.''
We also use YOMI-Bench to evaluate the reading abilities of representative LLMs.

The contributions of this study include:
\begin{itemize}
    \item We construct a challenging benchmark involving seven tasks that requires correct understanding of phonological readings in Japanese and release them on GitHub~\footnote{\scriptsize\url{https://github.com/benchmark-release/YOMI-Bench}} as publicly available linguistic resources.
    \item Using the benchmark, we evaluate one multilingual LLM, four Japanese-specific LLMs, and five commercial models, providing baselines for kanji reading performance.
\end{itemize}

\section{Related Work}
\paragraph{Grapheme-to-Phoneme (G2P) Benchmarks}
Grapheme-to-Phoneme (G2P) is a task that estimates a phoneme sequence from a given text.
Its main applications include speech synthesis and speech recognition.
Representative evaluation benchmarks for G2P include those based on the English-centered CMU Pronouncing Dictionary\footnote{\scriptsize\url{http://www.speech.cs.cmu.edu/cgi-bin/cmudict}}, as well as multilingual evaluation datasets released through shared tasks such as SIGMORPHON~\cite{cotterell2018conll}.
These benchmarks have been widely used to evaluate G2P systems, particularly for languages with alphabetic writing systems.

However, existing benchmarks primarily target languages with alphabetic writing systems, such as English, and are often designed under the assumption that the correspondence between written forms and sounds is relatively regular.
Therefore, it is not straightforward to directly apply these evaluation settings to languages such as Japanese and Chinese, where the correspondence between characters and pronunciations is often ambiguous and context-dependent.
Furthermore, most existing G2P benchmarks are designed for speech processing models and do not aim to directly evaluate the reading ability of text-based models such as LLMs.
To address this limitation, we construct a benchmark for evaluating Japanese reading ability in LLMs.

\paragraph{LLM Benchmarks in Japanese}
Several LLM evaluation benchmarks targeting Japanese have been released to date~\cite{saito2025buildlocallargelanguage}.
These benchmarks mainly focus on question-answering (QA) tasks, evaluating responses to content ranging from Wikipedia to more specialized domains.
A representative example is JMMLU~\cite{yin-etal-2024-respect}, which is based on MMLU and consists of datasets that have been translated and adapted to reflect Japanese cultural contexts.

In contrast to benchmarks that evaluate general Japanese language understanding, benchmarks related to reading have also been proposed.
The kana-to-kanji conversion benchmark AJIMEE-Bench~\footnote{\scriptsize\url{https://github.com/azooKey/AJIMEE-Bench}} has not been published as an academic paper, but it proposes tasks that convert kana characters into kanji based on their phonological readings, thereby involving reading-related processing.
However, none of these datasets explicitly evaluate reading ability.
However, in non-Roman-script languages, pronunciations often vary depending on context, making reading an important aspect of language understanding.
Accordingly, we construct a dataset to evaluate Japanese reading ability.

\begin{table}[tb]
\begin{center}
  \scalebox{0.8}{
  \label{tab:task_info}
  \begin{tabular}{l|r|r} \hline \hline
  Task Name & Task Format & Size\\\hline
  Kanji Reading QA (Single) & Binary QA & 240\\
  Kanji Reading QA (Multiple) & Binary QA & 450\\\hline
  Kanji Reading Prediction (Single) & Generation & 120\\
  Kanji Reading Prediction (Multiple) &Generation & 120\\\hline
  Rhyme Selection (Kanji)& Multiple QA & 120\\
  Rhyme Selection (Hiragana)& Multiple QA & 120\\\hline
  Rhyme Generation (Hiragana) & Generation & 120\\
  \hline \hline
  \end{tabular}
  }
  \caption{YOMI-Bench dataset statistics.}
  \vspace{-1.6em}
  \end{center}
\end{table}

\section{YOMI-Bench}
YOMI-Bench consists primarily of two types of tasks: tasks that evaluate kanji reading knowledge and tasks that require phonological understanding.

\subsection{Kanji Data Collection}
\label{sec:data}
The kanji characters targeted in the benchmark are selected based on the Jōyō Kanji List, a set of 2,136 officially designated kanji characters published by the Agency for Cultural Affairs of Japan.\footnote{\scriptsize\url{https://www.bunka.go.jp/kokugo_nihongo/sisaku/joho/joho/kijun/naikaku/kanji/joyokanjisakuin/index.html}}
The Jōyō Kanji List contains a total of 2,136 kanji characters, of which 803 have a single reading and 1,333 have two or more possible readings.
From this list, we randomly sample 100 kanji characters with a single reading and 100 kanji characters with multiple readings.

\subsection{Multi-Prompt Evaluation}
To mitigate biases arising from reliance on a single prompt~\cite{wang2024llmsperformmcqaselecting, chujie2024multiplechoice}, we adopt a multi-prompt evaluation strategy.
For each task, we design a base prompt and generate four paraphrased variants using ChatGPT while preserving the original semantic intent.
During evaluation, models are tested with five prompts, and the final score is computed as the average of their results.

\subsection{YOMI-Bench Tasks}
\paragraph{Kanji Reading Binary QA Task}
In this task, the model determines whether a given reading is a correct pronunciation of a kanji by answering Yes or No.
For example, the model is asked a question such as: ``Is \textit{kaku} the correct reading of the kanji 覚 in the word 覚醒?'' and must respond with either Yes or No.
This task evaluates whether an LLM possesses knowledge of kanji readings.

In this task, Yes/No-style prompts are constructed based on the kanji dataset.
Negative examples are generated by selecting readings from the set of readings in the kanji dataset that have the same number of readings as the target kanji and assigning them as incorrect readings.
In addition, each prompt is evaluated using a 4-shot setting, where two positive examples and two negative examples are provided. The positive and negative examples are constructed by sampling four instances from the test dataset for each.

\paragraph{Kanji Reading Prediction Task}
The Kanji Reading Prediction Task requires models to output the correct reading for a given Japanese word composed of kanji characters.
For example, given the input ``Please provide the reading of the word 覚醒,'' the expected output is \textit{kakusei}.

In this task, the predicted reading is extracted from the LLM output using regular expressions.
A score of 1 is assigned if the extracted reading exactly matches the gold-standard reading, and 0 otherwise; accuracy is computed as the average score.
For kanji with multiple readings, questions derived from the same kanji are treated as a single group, and the final score is computed by averaging the group-level accuracies.

\paragraph{Rhyme Selection Task}
The Rhyme Selection Task requires selecting a word that shares the same vowel sequence (i.e., forms a rhyme) with a given input word.
For example, given the prompt:
``Please select exactly one string from the following options that has the same vowel sequence as 覚醒 (\textit{kakusei}*).
(A) 論文, (B) 学生, (C) 委員, (D) 博士,''
the correct answer is ``(B) 学生 (\textit{gakusei}*).''\footnote{The readings of the options
 are: (A) 論文 (\textit{ronbun}* / Paper), (B) 学生 (\textit{gakusei}* / Student), (C) 委員会 (\textit{iinkai}* / Committee), and (D) 博士 (\textit{hakase}* / Doctor).}

Since the Rhyme Selection Task is formulated as a multiple-choice problem, we extract one of the options from (A) to (D) from the LLM's output.
If the extracted option matches the correct answer, a score of 1 is assigned; otherwise, a score of 0 is given.
The final accuracy is calculated by dividing the total score by the number of evaluation instances.
The target words are restricted to those containing kanji characters with a single possible reading, and we construct paired data consisting of one positive example and three negative examples.

\paragraph{Rhyme Generation Task}
The Rhyme Generation Task requires models to generate a hiragana word that shares the same vowel sequence with a given hiragana input word.
For example, given the prompt ``Please provide a word whose vowel sequence exactly matches that of かくせい (\textit{kakusei}),'' the expected output is はつめい (\textit{hatsumei}).

In the Rhyme Generation Task, generated rhymes are evaluated by comparing vowel sequences.
The rhyme is extracted from the model output using regular expressions and converted into a vowel sequence using the conversion table proposed by \citet{mibayashi2025rhyme}.
A score of 1 is assigned if the resulting sequence exactly matches the gold-standard vowel sequence, and 0 otherwise; accuracy is computed as the average score.

\begin{table*}[tb]
\begin{center}
  \scalebox{0.82}{
  \begin{tabular}{l|c|c|c|c|c|c|c}
    \hline \hline
    \multirow{2}{*}{Models}
      & \multicolumn{2}{c|}{Kanji Reading QA}
      & \multicolumn{2}{c|}{Kanji Reading Prediction}
      & \multicolumn{2}{c|}{Rhyme Selection}
      & \multirow{1}{*}{Rhyme Generation} \\
    \cline{2-8}
    \cline{6-6}
      & \multicolumn{1}{c|}{Single}
      & \multicolumn{1}{c|}{Multiple}
      & \multicolumn{1}{c|}{Single}
      & \multicolumn{1}{c|}{Multiple}
      & \multicolumn{1}{c|}{Kanji}
      & \multicolumn{1}{c|}{Hiragana}
      & \multicolumn{1}{c}{Hiragana}\\
    \hline
    \textit{Ministral-8B}
     & 0.6640 & 0.6355 & 0.4679 & 0.3023 & 0.2960 & 0.3400 & 0.6519 \\
    \hline
    \textit{Llama-3.1-Swallow-8B}
     & 0.7620 & 0.7782  & 0.8219 & 0.5116 & 0.2960 & 0.5020 & 0.8480 \\
    \textit{Llama-3-ELYZA-JP-8B}
     & 0.6980 & 0.6795 & 0.7900 & 0.4380 & 0.3040 & 0.5800 & 0.5439 \\
    \textit{llm-jp-3-7.2b}
     & 0.7350 & 0.6755 & 0.7999 & 0.5174 & 0.2279 & 0.2620 & 0.1660 \\
    \textit{llm-jp-3-13b}
     & 0.9620 & 0.8928 & 0.8960 & 0.5558 & 0.2780 & 0.3379 & 0.0980 \\
    \hline
    \textit{claude}
     & 0.9140 & 0.7733  & 0.9960 & 0.9216 & 0.9559 & 0.9840 & 0.7280 \\
    \textit{mistral}
     & 0.9960 & 0.8737  & 0.8019 & 0.5410 & 0.3460 & 0.5740 & 0.4679 \\
    \textit{gemini}
     & 0.9980 & 0.9817  & 0.9820 & 0.8631 & 0.9380 & 0.9480 & 0.5920 \\
    \textit{gpt-4o}
     & 1.0000 & 0.9404  & 0.9620 & 0.6943 & 0.5820 & 0.2580 & 0.5980 \\
    \textit{gpt-5}
     & 0.9970 & 0.9848  & 0.9840 & 0.9678 & 1.0000 & 0.9980 & 0.7800 \\
    \hline \hline
  \end{tabular}
  }
  \caption{The average quantitative evaluation results on YOMI-Bench.}
  \label{tab:quantitative_evaluation}
\end{center}
\vspace{-1em}
\end{table*}

\section{Baseline Experiments}
\subsection{Overview}
To demonstrate the difficulty and effectiveness of YOMI-Bench as a benchmark, we conduct a baseline evaluation.
In this study, we conduct evaluations using a total of ten models: one multilingual open LLM that supports Japanese, four Japanese-specific open LLMs, and five commercial models.
The specific models used in the evaluation are as follows:
\textit{Ministral-8B-Instruct-2410},
\textit{Llama-3.1-Swallow-8B-Instruct-v0.5}~\footnote{\scriptsize\url{https://huggingface.co/tokyotech-llm/Llama-3.1-Swallow-8B-Instruct-v0.5}},
\textit{Llama-3-ELYZA-JP-8B}~\footnote{\scriptsize\url{https://huggingface.co/elyza/Llama-3-ELYZA-JP-8B}},
\textit{llm-jp-3-7.2b-instruct3}~\footnote{\scriptsize\url{https://huggingface.co/llm-jp/llm-jp-3-7.2b-instruct3}}, and
\textit{llm-jp-3-13b-instruct3}~\footnote{\scriptsize\url{https://huggingface.co/llm-jp/llm-jp-3-13b-instruct3}}.
\textit{Ministral-8B-Instruct-2410} is a multilingual model, while the remaining four models are specialized for Japanese.
The following five models are used as commercial models in our evaluation:
\textit{claude-sonnet-4-5-20250929},
\textit{mistral-medium-2508},
\textit{gemini-2.5-flash},
\textit{gpt-4o}, and
\textit{gpt-5}.

The evaluation results for each task in our benchmark are shown in 
Table~\ref{tab:quantitative_evaluation}.
Overall, across all tasks, commercial models exhibit higher reading performance than Japanese-specific open models, despite not being explicitly specialized for Japanese.

\subsection{Discussion on Each Task}
\paragraph{Kanji Reading Binary QA Task}
As shown in Table~\ref{tab:quantitative_evaluation}, all Japanese-specialized models except \textit{llm-jp-3-13b} achieved performance of around 0.7.
Most models performed worse under multiple conditions than under a single condition.
This suggests that while models memorize single reading for kanji characters to some extent, they have difficulty predicting the correct reading according to the context.
Furthermore, the multilingual model also did not achieve high performance.
Since its performance is comparable to that of Japanese-specialized models, this result suggests that training with Japanese-specialized corpora alone may not necessarily lead to improved performance on kanji reading tasks.

\paragraph{Kanji Reading Prediction Task}
For kanji characters with a single reading, both Japanese-specialized LLMs and commercial LLMs showed high accuracy (more than 0.79).
These results confirm that LLMs are generally able to generate correct readings for kanji characters with a single reading.
In contrast, the multilingual LLM including Japanese showed a relatively low accuracy (0.4679).
These results suggest that differences in the amount of Japanese training data may influence kanji reading performance.
Furthermore, within the llm-jp series, which consists of models that share the same architecture but differ in parameter size, the 13B model outperformed the 7.2B model, suggesting that increasing the number of parameters may lead to improved model performance.

\paragraph{Rhyme Selection Task}
In this task, Japanese-specific models showed low performance overall.
Notably, even models that achieve high performance on the Kanji Reading task, such as \textit{llm-jp-13b} and \textit{Llama-3.1-Swallow-8B}, exhibit low accuracy on rhyme selection.
This pattern is also seen in commercial models such as \textit{mistral} and \textit{gpt-4o}.
Although these models appear to know correct kanji readings, they do not perform well in both kanji and hiragana settings, suggesting that knowledge of readings is not necessarily linked to their use.

\paragraph{Rhyme Generation Task}
Finally, in the Rhyme Generation Task, overall performance is low, and in particular, Japanese-specific open models are found to be almost incapable of generating rhymes.
Examples of outputs produced by Japanese-specific open models include generating ``やく (\textit{yaku})'' for the input ``かし (\textit{kasi})'', or ``せっけん (\textit{sekken})'' for the input ``けん (\textit{ken})'', indicating that these models fail to generate outputs with matching vowel sequences.
Even commercial models sometimes generate unnatural words.
For example, given the input ``せつな (\textit{setsuna})'', a commercial model generates ``めぐや (\textit{meguya})''.
Although the vowel sequences match in these cases, the generated outputs cannot be considered natural words.

\section{Conclusion}
We constructed YOMI-Bench, a benchmark specialized for evaluating kanji reading ability in Japanese LLMs.
Baseline experiments showed that even Japanese-specific models show low performance, and that commercial models also perform poorly on generation tasks that require consideration of kanji readings.
YOMI-Bench highlights the need for improving the reading-related abilities of LLMs.

\section*{Limitations}
This study focuses on evaluating the kanji reading ability of large language models in Japanese, and the findings may not directly generalize to other languages.
Our benchmark is constructed using the Jōyō Kanji set and common vocabulary, and therefore does not cover rare characters or domain-specific terms.

\section*{Ethical Considerations}
This work does not involve human subjects, personal data, or sensitive content.

\bibliography{custom}




\end{document}